\documentclass{article}
\usepackage{amsmath,epsfig}
\usepackage[preprint]{spconfa4}
\usepackage{algorithm}
\usepackage{algorithmic}
\usepackage{graphicx}
\usepackage{subfigure}
\usepackage{diagbox}
\usepackage{multirow}
\usepackage{booktabs}
\usepackage{amsmath}
\usepackage{booktabs}

\let\OLDthebibliography\thebibliography
\renewcommand\thebibliography[1]{
  \OLDthebibliography{#1}
  \setlength{\parskip}{0pt}
  \setlength{\itemsep}{0pt plus 0.3ex}
}

\begin{document}\sloppy

\def\x{{\mathbf x}}
\def\L{{\cal L}}

\title{Interpret the Predictions of Deep Networks via Re-Label Distillation}
%
\name{Yingying Hua$^{1,2}$, Shiming Ge$^{1,2,\ast}$
\thanks{This work was partially supported by grants from the National Key Research and Development Plan (2020AAA0140001), National Natural Science Foundation of China (61772513), Beijing Natural Science Foundation (19L2040), and the project from Beijing Municipal Science and Technology Commission (Z191100007119002). Shiming Ge is also supported by the Youth Innovation Promotion Association, Chinese Academy of Sciences.}
\thanks{$^{\ast}$Shiming Ge is the corresponding author.},
      Daichi Zhang$^{1,2}$}

\address{$^{1}$Institute of Information Engineering, Chinese Academy of Sciences, Beijing 100195, China
\\$^{2}$School of Cyber Security, University of Chinese Academy of Sciences, Beijing 100049, China
\\\{huayingying, geshiming, zhangdaichi\}@iie.ac.cn}

\maketitle

\begin{abstract}
Interpreting the predictions of a black-box deep network can facilitate the reliability of its deployment. In this work, we propose a re-label distillation approach to learn a direct map from the input to the prediction in a self-supervision manner. The image is projected into a VAE subspace to generate some synthetic images by randomly perturbing its latent vector. Then, these synthetic images can be annotated into one of two classes by identifying whether their labels shift. After that, using the labels annotated by the deep network as teacher, a linear student model is trained to approximate the annotations by mapping these synthetic images to the classes. In this manner, these re-labeled synthetic images can well describe the local classification mechanism of the deep network, and the learned student can provide a more intuitive explanation towards the predictions. Extensive experiments verify the effectiveness of our approach qualitatively and quantitatively.
\end{abstract}
\begin{keywords}
Deep neural network, interpretability, knowledge distillation
\end{keywords}
\section{Introduction}
Deep neural networks (DNNs) have proven their excellent capabilities on a variety of tasks.
The complex nonlinearity of deep models promotes extremely high accuracy, but also leads to the opacity and incomprehensibility~\cite{Khademi2018}.
In particular, it is hard to understand and reason the predictions of DNNs from the perspective of human.
These black-box models could cause serious security issues, such as the inability to effectively distinguish and track some errors, which diminishes the credibility of them. Therefore, it is of great significance to understand the decision-making process of DNNs and improve their interpretability to users.

The interpretability of deep models has already attracted an increasing attention in recent years~\cite{arrieta2020explainable}.
According to the scope of interpretability, it can be divided into global interpretability and local interpretability~\cite{molnar2020interpretable}. Global interpretability is based on the relationship between dependent and predictor variables to understand the predictions in the entire data set, that is, to establish the relationship between the output and input of deep models~\cite{Ras2018}. While local interpretability focuses on a single point and the local sub-region in the feature space around this point, and tries to understand the predictions based on the local region. Usually, local interpretability and global interpretability are used together to jointly explain the decision-making process of deep networks.

\begin{figure}[t]
	\centering
	\includegraphics[width=1.0\linewidth]{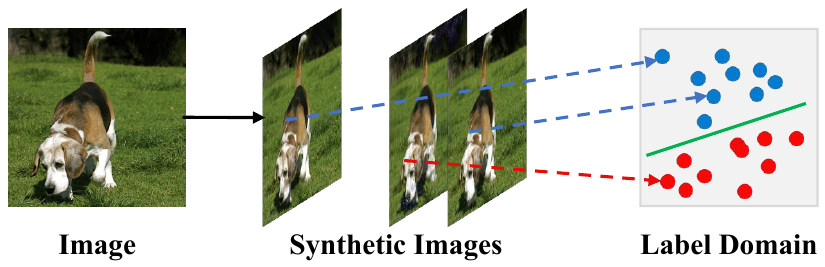}
	\caption{Motivation of our approach. To interpret the prediction of a specific image, we generate some synthetic images to deconstruct the hidden knowledge of DNNs. And the synthetic images can be well projected on both sides of the classification boundary in the label domain, which can represent the classification knowledge to interpret its prediction.}
	\label{fig1}
\end{figure}

\begin{figure*}[t]
    \centering
	\includegraphics[width=1.0\linewidth]{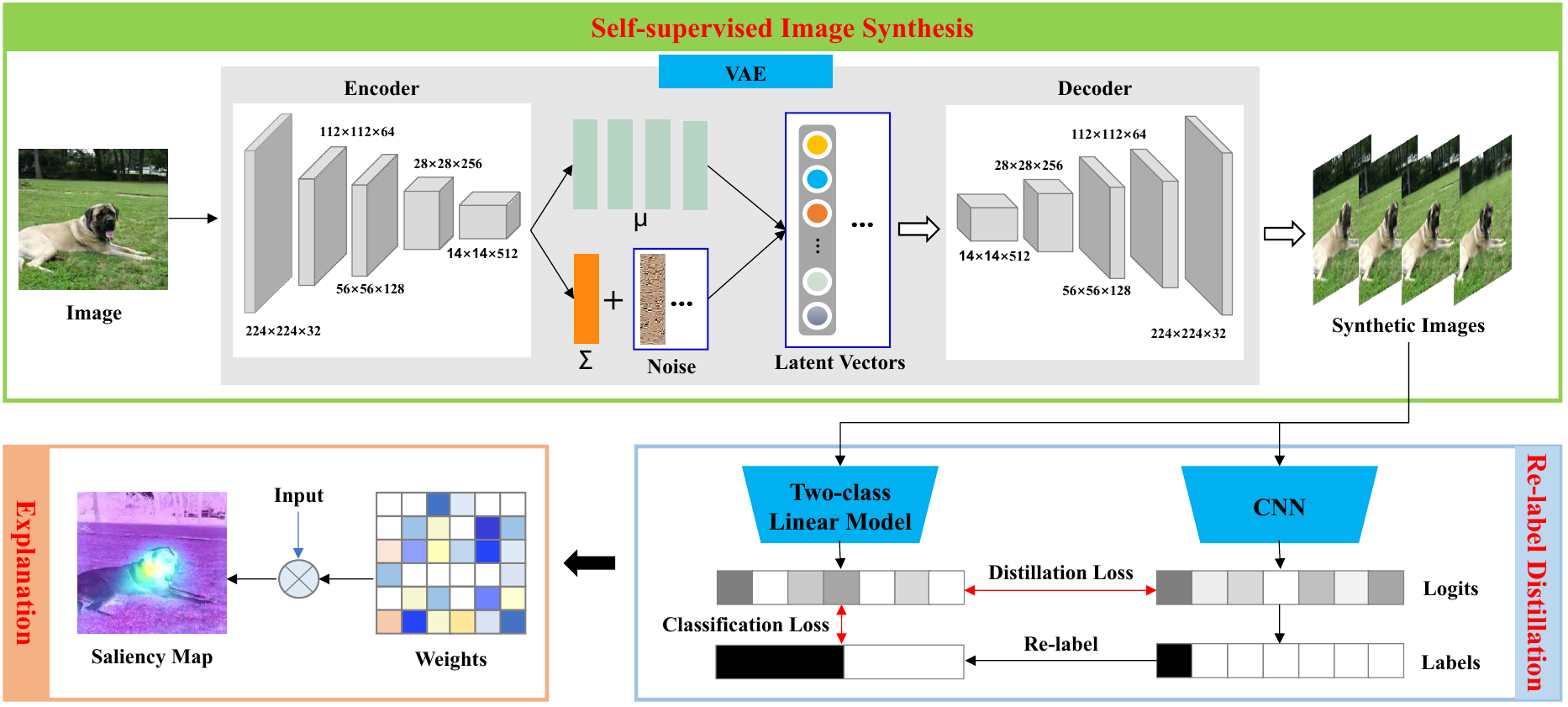}
	\caption{Overview of our approach. For a given image, we first use a pre-trained VAE to generate some synthetic images by perturbing the latent vector with random noise. Then, we re-label these synthetic images through a pre-trained CNN into one of two classed by identifying whether their predictions shift.
Finally, we train a two-class linear model by distilling the soft logits from CNN with these re-labeled synthetic images.
Therefore, the weights of the trained linear model can mark the location of the important features contributed to its prediction, which could generate a saliency map to interpret the prediction of the image.}
	\label{method}
\end{figure*}

In order to address this issue, many methods have been proposed and validated to meet this need for interpretability.
In terms of the interpretability, there are three main areas based on the purpose of these studies: 1) the first aims to make the components of deep networks more transparent, which is mainly achieved through visualization technology~\cite{Zhang2018}; 2) the second is achieved by learning a semantic graph~\cite{bass2020icam}. The implicit knowledge of these deep models can be characterized in an interpretable manner; 3) the last one is to generate a post-hoc explanation. For example, we can manipulate some interpretable models to explain the predictions afterwards~\cite{goode2020explaining}.
Nonetheless, these traditional methods~\cite{zhang2020a} still need some improvements, such as the effect of explanation does not ensure that human can totally understand deep networks, and the universality of these explanation methods is limited by the specific model structure.
More importantly, a pre-trained deep network still can not explain the prediction of a specific image.
This is a crucial issue to interpret the predictions of deep networks for image classification.

To this end, we propose a re-label distillation approach to interpret the prediction of deep networks.
As shown in Fig.~\ref{fig1}, we first generate some synthetic images to characterize the feature distribution of the image, and then re-label these synthetic images through a pre-trained DNN to capture the local classification knowledge near the image~\cite{Wang2019}. In this way, we can transfer the decision boundary knowledge into these re-labeled synthetic images.
The interpretation of the prediction can be easily achieved by learning the deconstructed boundary knowledge in an understandable way.
Based on this, we train an interpretable model in low-dimensional space with these re-labeled images by distilling the boundary knowledge of DNN~\cite{Liu2019}.
The trained student model learns a direct map from the image to the prediction, which can mark the important features contributed to its prediction. Therefore, the proposed re-label distillation approach can interpret the predictions of deep networks.
To validate the effectiveness of our approach, we conduct experiments through qualitative and quantitative evaluations and our re-label distillation approach performs impressively for explanations.

To summarize, the main contributions are as follows: 1) This paper proposes a re-label distillation approach to interpret the predictions of deep networks by distilling into an interpretable model in low-dimensional space; 2) The proposed approach designs an algorithm to train a two-class linear model for explanation. We first generate some synthetic images to represent the classification knowledge of DNN with a VAE. And then we re-label them by identifying whether their predictions shift, which could transfer the boundary knowledge of DNN into them. Finally, we train a two-class linear model through distillation with these re-labeled synthetic images; 3) The experimental results verify the effectiveness of our proposed interpretable approach qualitatively and quantitatively.

\section{Approach}
The overview of our interpretable approach is illustrated in Fig.~\ref{method}.
In the following subsections, we will analyse the proposed re-label distillation approach in details.

\subsection{Problem Formulation}
Inspired by transfer learning, we can not only transfer the model structure, but also transfer the interpretability of deep networks.
Through some interpretable models, such as linear models or decision trees, deep networks can be reconstructed by distilling their hidden knowledge into these interpretable models.
We train an interpretable model to learn the output of the black-box model, so that we can establish an interpretable relationship between the input and output of the black-box model and achieve the interpretability of the predictions.

However, there are two fundamental problems: 1) The nonlinearity of DNN makes it hard to find a perfect interpretable model to capture the entire classification mechanism; 2) Traditional knowledge distillation leads to the loss of a lot of effective information and seriously affects the performance of the student. In order to solve the problems, we generate some synthetic images to represent the classification knowledge of deep networks, and then transfer the local boundary knowledge of the input into an interpretable model. Therefore, we could understand and reason the predictions of DNN.

The foundation of our interpretable approach is mainly to train an interpretable student model $\mathcal S$ to interpret the prediction of a pre-trained DNN $\mathcal T$ with respect to a given image $x$.
The student model can learn a direct map $m$ from the image to the prediction $y$,
\begin{equation}\label{eq0}
  y = x\cdot m,
\end{equation}
where $m$ can mark the location of the effective features contributed to its prediction.
Therefore, we use the parameters of the student as the weight of the features contributed to its prediction, which can be achieved by learning a student to imitate the DNN.
In this way, we need some synthetic images $x'$ to capture the informative classification knowledge towards the DNN. A generator can reconstruct the image to generate some synthetic images for the training of the student.

To formalize the idea of matching the outputs between the student model and the teacher model, we minimize an objective function to match the probability distributions and the outputs $y'$ of the student with that of the teacher. The loss function for the training of student $\mathcal S$ can be defined as,
\begin{align}\label{eq1}
\arg\min_{w} \mathcal L(P_{\mathcal S}(x';w), P_{\mathcal T}(x'))
   + \mathcal L(P_{\mathcal S}(x';w), y'),
\end{align}
where $\mathcal L(\cdot)$ is the cross-entropy classification loss, $P_{\mathcal S}(x';w)$ denotes the student's prediction distribution with the training parameter $w$, and$P_{\mathcal T}(x')$ denotes the teacher's distribution.
Therefore, we can train an interpretable model with some synthetic images to interpret the predictions of deep networks.

\subsection{Self-supervised Image Synthesis}
In our interpretable framework, the role of the synthetic images is to characterize the classification mechanism of the deep network. The clarity of the synthetic images is not our focus, and we prefer to generate semantically meaningful images of different categories. So we use variational autoencoder (VAE)~\cite{kusner2017grammar} as this generator.
A VAE comprises two sub-networks of an encoder $\boldsymbol{p}$ and a decoder $\boldsymbol{q}$.
The aim of encoder is to learn a latent representation that describes each latent attribute in probabilistic terms.
To generate the synthetic images $X=\{x_i'\}_{i=1}^n$, we add some random noise $\boldsymbol{\varepsilon}_i$ to the latent vector,
\begin{equation}\label{eq2}
  \boldsymbol{z}_i  = \mu + \boldsymbol{\varepsilon}_i * \Sigma,
\end{equation}
where $\mu$ and $\Sigma$ are the mean and standard deviation towards the learned representation of the encoder.
The input of our decoder model can be generated by randomly sampling from each latent representation, and the reconstructed output of the decoder $\boldsymbol{q}$ is,
\begin{equation}\label{eq3}
  x_i' = \boldsymbol{q}_\sigma(x|\boldsymbol{z}_i),
\end{equation}
where $\sigma$ is the training parameter of the decoder $\boldsymbol{q}$.

When training a VAE, we expect the decoder to be able to reconstruct the given input as accurately as possible for any sampling of the latent distributions.
Therefore, the values that are close to each other in the latent space should correspond to very similar reconstructions.
The training process of a VAE can be formulated as,
\begin{equation}\label{eq4}
  \arg\min_{\sigma,\varphi} \sum_{i=1}^n \mathcal L(x, x_i';\sigma,\varphi) + \mathcal D(\boldsymbol{p}(\boldsymbol{z}_i|x;\varphi), \boldsymbol{q}(\boldsymbol{z}_i;\sigma)),
\end{equation}
where $\varphi$ is the parameter of the encoder, $\mathcal D(\cdot)$ measures the similarity between the learned latent distribution $\boldsymbol{p}(\boldsymbol{z}_i|x;\varphi)$ and the true prior distribution $\boldsymbol{q}(\boldsymbol{z}_i;\sigma)$.
The training of VAE is regularised to avoid over-fitting and ensure that the latent space has good properties to enable generative process.

\begin{figure}[t]
	\centering
	\includegraphics[width=1.0\linewidth]{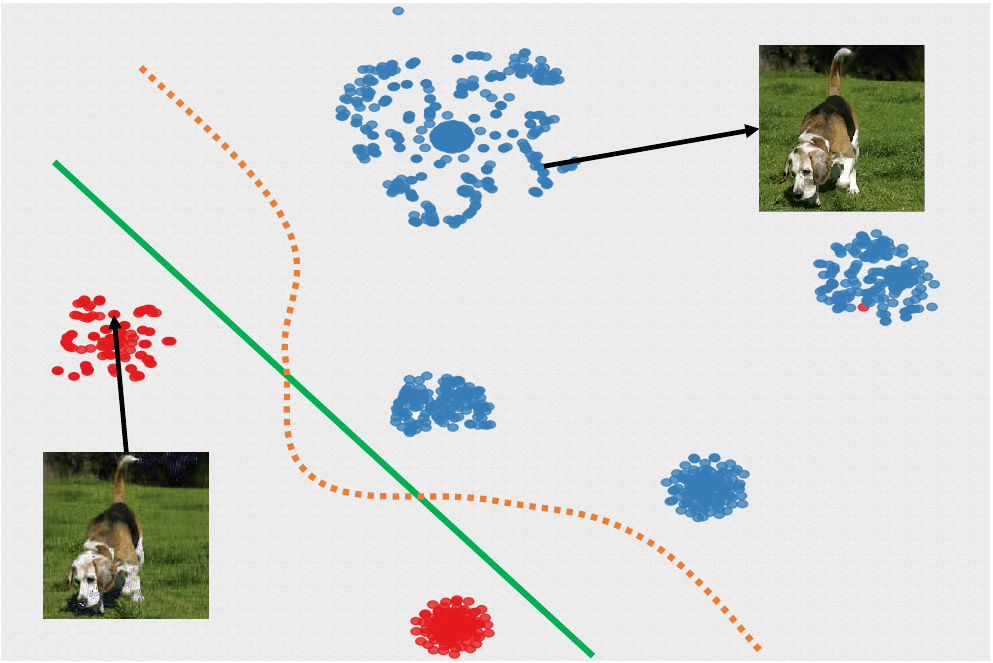}
	\caption{An example of t-SNE~\cite{JMLR:v9:vandermaaten08a} plots. Points are colored by their reconstructed labels. The plots indicate that the synthetic images are sortable and could be projected into the label domain to represent the boundary knowledge of deep networks.
}
	\label{sne}
\end{figure}

\begin{figure*}[htb]
	\centering
	\includegraphics[width=1.0\linewidth]{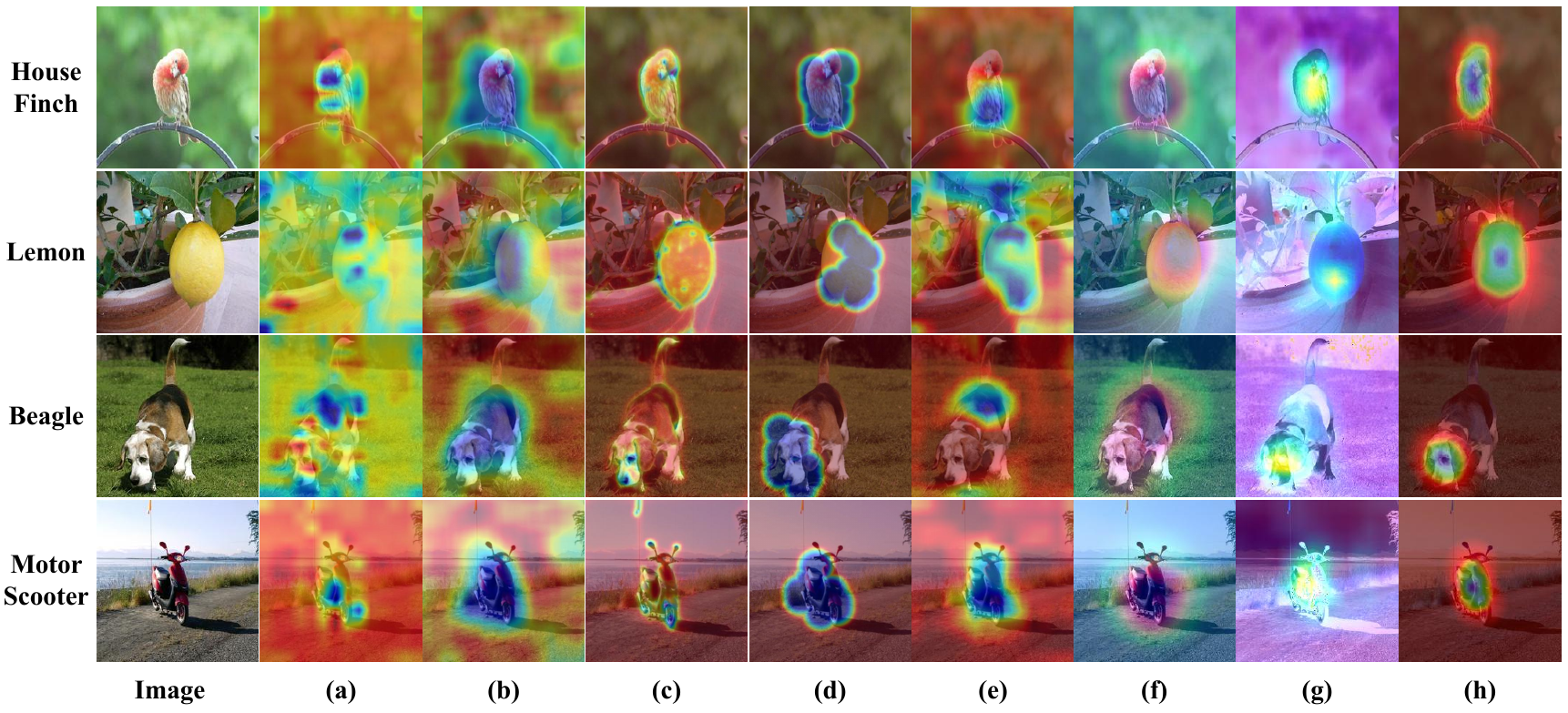}
	\caption{The qualitative comparisons with state-of-the-art methods, including (a) Linear approximation, (b) RISE~\cite{petsiuk2018rise}, (c) Excitation backprop~\cite{2018Top}, (d) Extremal perturbations~\cite{fong2019understanding}, (e) Grad-CAM~\cite{selvaraju2020grad}, (f) Score-CAM~\cite{2020Score}, (g) Occlusion sensitivity~\cite{10.1007/978-3-319-10590-1_53}, (h) Re-Label distillation (ours).
Our saliency maps mark the target area of the image more accurately and show the degree of importance to the predictions more clearly, which means our method generates a better visual explanation than others.}
	\label{vis}
\end{figure*}
\subsection{Re-label Distillation}
In order to understand and reason the predictions of deep networks, we use knowledge distillation~\cite{hinton2015distilling} to train an interpretable model by deconstructing the hidden knowledge inside the teacher, which could achieve an interpretable prediction for a given image.
Based on the above analysis, we apply a VAE to generate some synthetic images regarding a specific image and transfer the classification knowledge of deep networks into them. Therefore, we propose the re-label distillation approach to interpret its prediction.

The mathematical description of the re-label distillation will be described below.
We feed these re-labeled synthetic images $\{x_i',y_i'\}$ to a pre-trained deep network $\mathcal T$ and re-label them into two class by identifying whether their predictions $f_{\mathcal T}(\cdot)$ shift. The predictions of these synthetic images from the teacher are referred as their true labels. In this way, these reconstructed labels $\{y_i'\}$ include the classification boundary knowledge of the deep network and can be seen as the relation of these predicted labels with respect to the synthetic images. And the synthetic images are re-labeled by
\begin{equation}\label{eq5}
\begin{aligned}
y_i' = \begin{cases}
1, & f_{\mathcal T}(x_i')=f_{\mathcal T}(x) \\
0, & f_{\mathcal T}(x_i')\neq f_{\mathcal T}(x) \\
\end{cases}.
\end{aligned}
\end{equation}
As shown in Fig.~\ref{sne}, these synthetic images can be classified based on the t-SNE~\cite{JMLR:v9:vandermaaten08a} plots on ResNet50, which means the classification boundary knowledge of DNN has been transferred to these re-labeled synthetic images.
Then, we train a two-class linear model with these re-labeled synthetic images by distilling the soft knowledge of the deep network.

The loss of our proposed re-label distillation to train the student model $\mathcal S$ can be defined as,
\begin{equation}\label{eq6}
\begin{aligned}
  L(w) & \!=\! \lambda_1\mathcal L_{soft}(x';w) + \lambda_2\mathcal L_{hard}(x',y';w)
             \\ & \!=\! \sum_{i=1}^n \lambda_1\|P_{\mathcal S}(x_i',w) \!-\! P_{\mathcal T}(x_i')\|_2
             \! + \! \lambda_2\|f_{\mathcal S}(x_i',w) \!-\! y_i'\|_1,
\end{aligned}
\end{equation}
where $w$ denotes the weights of the linear model, $\lambda_1$ and $\lambda_2$ are the weight coefficients, $f_{\mathcal S}(\cdot)$ denotes the prediction of the student model.
The trained linear model establishes an interpretable relation between the prediction and the input.
The weights $w$ could measure the significance of different pixels contributed to its prediction.
Therefore, we could obtain an explanation by locating the salient features onto the image.

\begin{table*}[t]
\centering
\caption{Quantitative comparisons on the deletion (lower is better) and insertion (higher is better) metrics.}
\begin{tabular}{c|c|cccccc}
  \hline
  Metrics & Methods & Grad-CAM~\cite{selvaraju2020grad} & Sliding Window~\cite{10.1007/978-3-319-10590-1_53} & LIME~\cite{Ribeiro2016} & RISE~\cite{petsiuk2018rise} & FGVis~\cite{wagner2019interpretable} & Our method\\
  \hline
  \multirow{2}{*}{Deletion} & ResNet50 & 0.1421 & 0.1232 & 0.1217 & 0.1076 & 0.0644 & \textbf{0.0627}$\downarrow$\\
   & VGG16 & 0.1158 & 0.1087 & 0.1014 & 0.0980 & 0.0636 & \textbf{0.0513}$\downarrow$\\
  \hline
  \multirow{2}{*}{Insertion} & ResNet50 & 0.6766 & 0.6618 & 0.6940 & \textbf{0.7267} & - & 0.7190\\
   & VGG16 & 0.6149 & 0.5917 & 0.6167 & 0.6663 & - & \textbf{0.7981}$\uparrow$\\
  \hline
\end{tabular}
\label{tab:result}
\end{table*}

\section{Experiments}
To evaluate our re-label distillation approach, we compare with 8 state-of-the-art interpretable approaches in 2 typical deep networks (ResNet50 and VGG16 trained on ImageNet), including RISE~\cite{petsiuk2018rise}, Excitation backprop~\cite{2018Top}, Extremal perturbations~\cite{fong2019understanding}, Grad-CAM~\cite{selvaraju2020grad}, Score-CAM~\cite{2020Score}, Occlusion sensitivity~\cite{10.1007/978-3-319-10590-1_53}, LIME~\cite{Ribeiro2016}, and FGVis~\cite{wagner2019interpretable}.

\subsection{Experiment Settings}
We provide a comprehensive evaluation for our approach qualitatively and quantitatively.
The images of ImageNet are preprocessed to 224 $\times$ 224 $\times$ 3 to train our interpretable framework.
We use a VAE of 500D latent space to generate 1000 synthetic images for each input by perturbing the latent vector with some random noise.
And the coefficients of the re-label distillation loss $\lambda_1$ and $\lambda_2$ are set to 0.7 and 0.3 in Eq.(\ref{eq6}).
Qualitative evaluation is mainly carried out through the visualization of saliency maps, which present the important features contributed to the predictions. And quantitative evaluation uses the metrics of deletion and insertion~\cite{petsiuk2018rise}.
The deletion metric measures the decrease of the prediction probability as more and more salient features are removed from the image, while the insertion measures the increase of the prediction probability as more and more features are inserted.

\subsection{Qualitative Results}
We could explain the decision-making process of deep networks by generating a saliency map, which colors each pixel according to its importance to the prediction.
Based on this, we use the weight parameters of the trained linear model to generate the feature-importance maps for qualitative evaluation.
As shown in Fig.~\ref{vis}, our re-label distillation approach can mark the significant regions contributed and ignore some irrelevant background information. And compared with other state-of-the-art methods, our saliency map could mark the more accurate target area. Although extremal perturbations~\cite{fong2019understanding} could generate an accurate saliency map, our method also shows the importance of these salient features more clearly. By visualizing the saliency map, we can observe the important pixels of the image contributed to its prediction and build an interpretable relationship between them.
Therefore, our proposed re-label distillation approach has a better performance in explaining the predictions of deep networks.

\begin{figure}[t]
  \centering
   \includegraphics[width=1.0\linewidth]{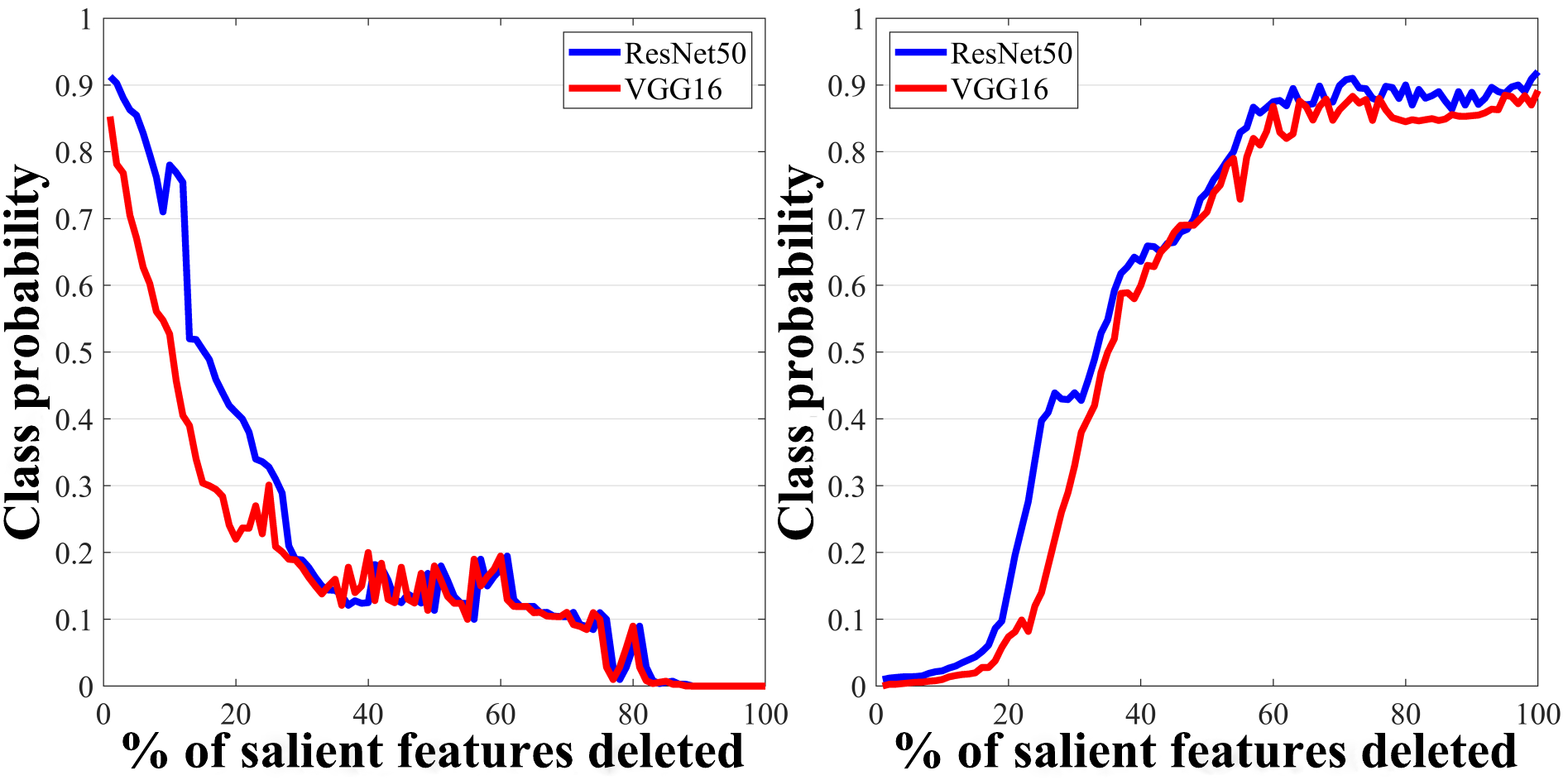}
  \caption{Quantitative results on ResNet50 and VGG16. With the deletion (left) or insertion (right) of the salient features, the obvious change in class probability validates the significance of these features contributed to the models' predictions.}
  \label{curve}
\end{figure}

\begin{figure}[t]
	\centering
     \includegraphics[width=1.0\linewidth]{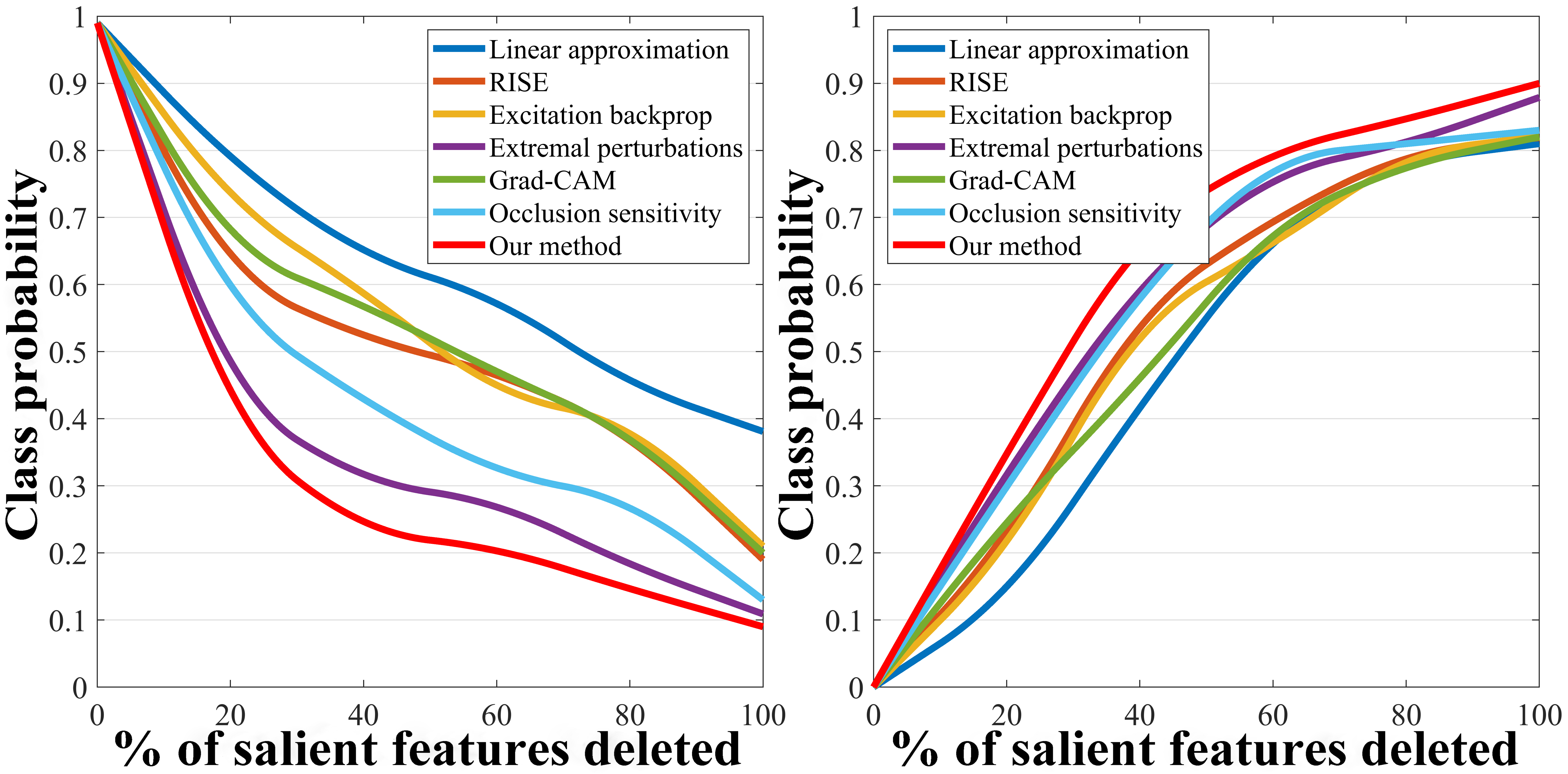}
	\caption{Comparison results with other interpretable methods. With the deletion (left) or insertion (right) of the salient features, our method increases or decreases more sharply in probability, which indicates a more accurate explanation.}
	\label{fig3}
\end{figure}
\subsection{Quantitative Results}
To conduct the quantitative experiments, we use the deletion and insertion metrics to evaluate the saliency maps generated with re-label distillation, and also compare the results with other interpretable approaches. As shown in Fig.~\ref{curve}, the obvious change of the class probability indicates the significance of the salient features towards the predictions of ResNet50 and VGG16. Fig.~\ref{fig3} presents the comparison results with other methods on ResNet50. From the graph above we can see that the class probability is more sensitive to the changes (deletion and insertion) of the salient features under our method.
Tab.~\ref{tab:result} provides the comparison results of the deletion and insertion metrics averaged on the ImageNet validation dataset. The baseline results are taken from~\cite{petsiuk2018rise}. We use the AUC (area under curve) to measure these two metrics, which means the lower deletion and the higher insertion represent a better explanation. The table above illustrates that our deletion metric is 0.0627 on ResNet50 and 0.0513 on VGG16, and our insertion metric is 0.7190 on ResNet50 and 0.7981 on VGG16. We have achieved significantly better results, and there is only a small gap compared with RISE~\cite{petsiuk2018rise} on the insertion of VGG16.
Our saliency map has a greater impact on the prediction probability from deep networks, which confirms that the re-label distillation approach could generate a more accurate explanation than others.
Therefore, the re-label distillation generally outperforms these interpretable methods.

\section{Conclusion}
In this paper, we propose a re-label distillation approach to interpret the predictions of deep networks.
We first apply a VAE to generate some synthetic images. 
Then, we feed these synthetic images into deep networks for the reconstructed labels which are annotated by identifying whether they shift. Finally, we train a two-class linear model on these re-labeled synthetic images by distilling the soft logits and the hard labels.
Therefore, the trained linear model learns a direct map to locate the salient features of the input towards its corresponding prediction.
The experiments demonstrate that our approach can interpret the predictions of deep networks better.
And this work also provides research ideas for the future development of explainable artificial intelligence.

\bibliographystyle{IEEEbib}
\bibliography{main}

\end{document}